\documentclass[sigconf, screen]{acmart}
\usepackage{multirow}
\usepackage{array,hhline}
\usepackage{graphicx}
\usepackage{framed}

\AtBeginDocument{%
  }

\setcopyright{acmcopyright}
\copyrightyear{2023}
\acmYear{2023}
\acmDOI{XXXXXXX.XXXXXXX}

\acmConference[ICMR'23]{International Conference on Multimedia Retrieval}{June 12--15,
  2023}{Thessaloniki, Greece}
\acmPrice{15.00}
\acmISBN{978-1-4503-XXXX-X/18/06}

\acmSubmissionID{4676}

\begin{document}

\title{A Unified Model for Video Understanding and Knowledge Embedding with Heterogeneous Knowledge Graph Dataset}

\author{Jiaxin Deng}
\authornote{Interns at MMU, KuaiShou Inc.}
\affiliation{%
  \institution{National Laboratory of Pattern Recognition, Institute of Automation, Chinese Academy of Sciences}
  \streetaddress{95 Zhongguancun East Road}
  \city{Haidian Qu}
  \state{Beijing Shi}
  \country{China}
  \postcode{100190}
}
\email{dengjiaxin2022@ia.ac.cn}

\author{Dong Shen}
\affiliation{%
  \institution{MMU KuaiShou Inc.}
  \streetaddress{6 Shangdi West Road}
  \city{Haidian Qu}
  \state{Beijing Shi}
  \country{China}
  \postcode{100085}
}
\email{shendong@kuaishou.com}

\author{Haojie Pan}
\affiliation{%
  \institution{MMU KuaiShou Inc.}
  \streetaddress{6 Shangdi West Road}
  \city{Haidian Qu}
  \state{Beijing Shi}
  \country{China}
  \postcode{100085}
}
\email{panhaojie@kuaishou.com}

\author{Xiangyu Wu}
\affiliation{%
  \institution{MMU KuaiShou Inc.}
  \streetaddress{6 Shangdi West Road}
  \city{Haidian Qu}
  \state{Beijing Shi}
  \country{China}
  \postcode{100085}
}
\email{wuxiangyu@kuaishou.com}

\author{Ximan Liu}
\affiliation{%
  \institution{MMU KuaiShou Inc.}
  \streetaddress{6 Shangdi West Road}
  \city{Haidian Qu}
  \state{Beijing Shi}
  \country{China}
  \postcode{100085}
}
\email{liuximan@kuaishou.com}

\author{Gaofeng Meng}
\authornote{Corresponding author.}
\affiliation{%
  \institution{National Laboratory of Pattern Recognition, Institute of Automation, Chinese Academy of Sciences}
  \streetaddress{95 Zhongguancun East Road}
  \city{Haidian Qu}
  \state{Beijing Shi}
  \country{China}
  \postcode{100190}
}
\email{gfmeng@nlpr.ia.ac.cn}

\author{Fan Yang}
\affiliation{%
  \institution{MMU KuaiShou Inc.}
  \streetaddress{6 Shangdi West Road}
  \city{Haidian Qu}
  \state{Beijing Shi}
  \country{China}
  \postcode{100085}
}
\email{yangfan@kuaishou.com}

\author{Size Li}
\affiliation{%
  \institution{MMU KuaiShou Inc.}
  \streetaddress{6 Shangdi West Road}
  \city{Haidian Qu}
  \state{Beijing Shi}
  \country{China}
  \postcode{100085}
}
\email{lisize@kuaishou.com}

\author{Ruiji Fu}
\affiliation{%
  \institution{MMU KuaiShou Inc.}
  \streetaddress{6 Shangdi West Road}
  \city{Haidian Qu}
  \state{Beijing Shi}
  \country{China}
  \postcode{100085}
}
\email{furuiji@kuaishou.com}

\author{Zhongyuan Wang}
\affiliation{%
  \institution{MMU KuaiShou Inc.}
  \streetaddress{6 Shangdi West Road}
  \city{Haidian Qu}
  \state{Beijing Shi}
  \country{China}
  \postcode{100085}
}
\email{wangzhongyuan@kuaishou.com}

\renewcommand{\shortauthors}{Deng et al.}

\begin{abstract}
  Video understanding is an important task in short video business platforms and it has a wide application in video recommendation and classification. Most of the existing video understanding works only focus on the information that appeared within the video content, including the video frames, audio and text. However, introducing common sense knowledge from the external Knowledge Graph (KG) dataset is essential for video understanding when referring to the content which is less relevant to the video. Owing to the lack of video knowledge graph dataset, the work which integrates video understanding and KG is rare. In this paper, we propose a heterogeneous dataset that contains the multi-modal video entity and fruitful common sense relations. This dataset also provides multiple novel video inference tasks like the Video-Relation-Tag (VRT) and Video-Relation-Video (VRV) tasks. Furthermore, based on this dataset, we propose an end-to-end model that jointly optimizes the video understanding objective with knowledge graph embedding, which can not only better inject factual knowledge into video understanding but also generate effective multi-modal entity embedding for KG. Comprehensive experiments indicate that combining video understanding embedding with factual knowledge benefits the content-based video retrieval performance. Moreover, it also helps the model generate better knowledge graph embedding which outperforms traditional KGE-based methods on VRT and VRV tasks with at least \textbf{42.36\%} and \textbf{17.73\%} improvement in HITS@10.
\end{abstract}




\maketitle

\section{Introduction}
With the rapid development of various information-sharing media on the Internet, short video platforms such as TikTok and KuaiShou have become the most ubiquitous way for entertainment and socializing. Hundreds of millions of videos are produced daily on such widespread platform. Hence, the quality of video understanding not only greatly affects the basic tasks such as video classification but also has a significant influence on the advanced application scenarios like video tagging or video recommendation. In this field, video inference means retrieving the relevant tail entity (tag or video) when given head entity (video) and relation.
\begin{figure}
\centering
\includegraphics[width=.40\textwidth]{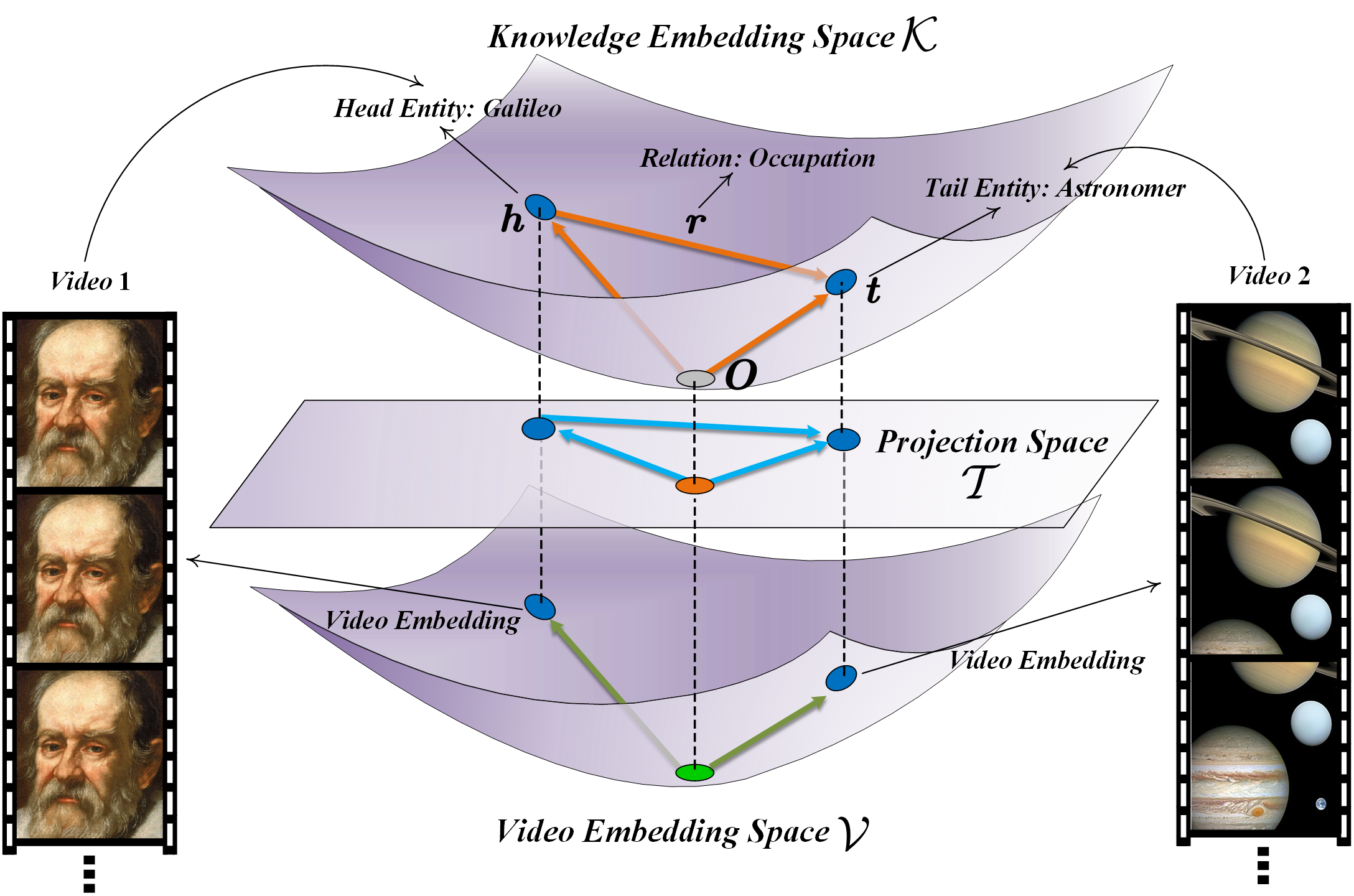}
\caption{\textbf{Example of heterogeneous embedding space with the triplet: Galileo-Occupation-Astronomer.} Video 1 and video 2 are tagged with the ``Galileo'' and ``Astronomer'' respectively and they are projected into the same space of tag embedding. Compared to the existing methods which only focus on the video content and tag-relation-tag inference task, the video-relation-tag, video-relation-video inference task and the content-based video retrieval task are all realizable in this heterogeneous embedding space.}
\label{fig1}
\vspace{-0.5cm}
\end{figure}
Traditional methods \cite{feichtenhofer2019slowfast} \cite{kang2006correlated} \cite{ren2015faster} only focus on retrieving content-based tag from the fusion of different modalities such as text, speech audio and video frames. For example, in Figure 1, the tag ``Galileo" is assigned to video 1 which narrates the life story of Galileo and ``Astronomer" is allied to video 2 which shows the concept of astronomer. However, these methods may fail to infer the details which are less relevant to the video content. Take video 1 for example, it is hard for the aforementioned methods to retrieve the tag ``Astronomer" because the content about astronomer is absent in video 1. Hence, the combination of Knowledge Graph (KG) and video understanding is essential to infer the cross-video knowledge. To be specific, in the above-mentioned example video 1 and video 2 can be matched to the entity ``Galileo" and ``Astronomer" which come from the external knowledge graph and the relation ``Occupation" provides a bridge between content-based and cross-video knowledge. The involvement of knowledge graph relation ``Occupation" is the key for inferencing the tag or video content which may not appear in the given video.

The existing Knowledge Graph Embedding (KGE) methods map the text entity and relation into a low-dimensional space and different distance matrix is designed for modeling the topology information of knowledge graph, e.g., TransE \cite{bordes2013translating}, TransH \cite{wang2014knowledge}, TransR \cite{lin2015learning} and RotatE \cite{sun2019rotate}. Nevertheless, these works cannot take full advantage of the abundant information in text. Aimed at solving the above problems, K-BERT \cite{liu2020k} combines the pre-trained language model with the KGE method which not only benefits various NLP applications but also enhances the embedding representation of KG. However, all the above-mentioned methods only focus on the text modality. Owing to the lack of video-based multi-modal knowledge graph dataset, the method which integrates video understanding and knowledge embedding is rare. The most related video understanding methods enhanced by KG are ACAR-Net \cite{pan2021actor}, PaStaNet \cite{li2020pastanet} and AKU \cite{ma2022visual}. However these works mainly focus on the fine-grained pattern in video like human activities recognition while our work mainly focuses on video classification and video tagging. Apart from that, the multi-modal knowledge graph dataset like MMEA \cite{chen2020mmea} and FreeBase \cite{bordes2013translating} has been proposed and widely studied in the task of entity alignment \cite{sun2018bootstrapping}, entity prediction and link prediction. However, the MMEA and FreeBase dataset only provide text entity and relevant images which are crawled from Internet search engines. These datasets lack multi-modal entities with rich semantic information, such as videos.

In order to combine video understanding with knowledge graph embedding, three challenges should be concerned. First, a video-based multi-modal knowledge graph dataset should be formed. Second, an effective embedding representation of video should be extracted and learned. The embedding representation of video should not only express the video individual properties, but also can be embedded into the same embedding space of knowledge graph embedding. Last but not least, videos contain rich multi-modal information from video frames, text and audio whereas knowledge graphs contain extensive structural factual knowledge, which is stored in the form of head entity-relation-tail entity triplet. So a unified model should be designed to tackle heterogeneity issue and fully exploit the combination of video content and knowledge triplets.

Considering the above three challenges, in this paper, we first propose a heterogeneous knowledge graph video dataset based on the Company short video platform. This dataset contains rich multi-modal entities and various common-sense relations. Based on this dataset, we propose a method that simultaneously integrates the multi-modal embedding representation of videos and the knowledge graph embeddings of triplets. To be specific, we first design a transformer architecture for video understanding and embedding extraction. This part can take full advantage of the abundant information from the multi-modal video and it is optimized with the video classification objective. Apart from that, in order to combine the video representation with factual knowledge, we design a video understanding and knowledge embedding integration model. By adopting the CLIP-based \cite{radford2021learning}  content retrieval objective, this model can project the video embeddings into the same semantic space as the knowledge graph embedding. Apart from that, it optimizes the video embedding and KGE objectives simultaneously. So our method not only helps KG embedding inherit the strong ability of video understanding but also enables the video embedding to conduct video inference tasks.

To summarize, our contributions are as follows:

\begin{itemize}
\item We define a novel formulation of the Video-Relation-Video and Video-Relation-Tag inference tasks based on the heterogeneous video knowledge graph dataset.
\item We propose a transformer architecture for multi-modal video understanding and knowledge graph embedding integration.
\item Extensive experiments indicate that our method achieves the \emph{state-of-the-art} performance on video inference tasks and it also brings improvement on content-based video retrieval tasks. 
\end{itemize}

\section{Related Work}
\begin{table*}[h]
\begin{center}
\caption{The meta information of the related dataset. $\mathcal{V}$, $\mathcal{A}$ and $\mathcal{T}$ represent video, audio and text respectively.}
\label{tab2}
\begin{tabular}{p{3cm}<{\raggedright}|p{1.5cm}<{\centering}|p{1.5cm}<{\centering}|p{1.5cm}<{\centering}|p{1.5cm}<{\centering}|p{2cm}<{\centering}|p{2cm}<{\centering}}
\hline
Dataset        & Entities & Triplets & Relations &Tags& Modalities & Videos     \\ \hline
Company-400M   & -        & -        & -              &427,249& \{$\mathcal{V}$,$\mathcal{A}$,$\mathcal{T}$\}  & 47,134,152 \\
Company-5M     & 248,324  & 832,577  & 5,150          &84,838& \{$\mathcal{V}$,$\mathcal{A}$,$\mathcal{T}$\}  & 5,714,531  \\
CN-DBpedia sub & 101,002  & 465,714  & 4,987          &-& \{$\mathcal{T}$\}      & -     \\ \hline
\end{tabular}
\end{center}
\tiny
\end{table*}
\subsection{Knowledge Graph Embeddings(KGE)}
Link and entity prediction in pure-text knowledge graph has been universally formulated with knowledge graph embedding methods recently. By mapping the entity and relation into a low-dimensional vector space $\mathbb{R}^{k}$, the main idea of these methods is to design a proper distance function $f_{r}(h, t)$ for the triplets which measures the reasonability of head entity \texttt{h} to tail entity \texttt{t} with the link relation \texttt{r}. For example, TransE \cite{bordes2013translating} defines the distance function as $f_{r}(h, t) = \|\mathbf{h}+\mathbf{r}-\mathbf{t}\|$ which hypothesizes that the relation should be a simple transition of the head entity and tail entity. TransE performs well on the 1-to-1 relations but fails on the 1-to-N, N-to-1 and N-to-N relations. Based on TransE, the subsequently proposed methods like TransH \cite{wang2014knowledge} and TransR \cite{lin2015learning} try to introduce the relation-specific entity embeddings to overcome the disadvantage of TransE and these methods gain performance improvement by increasing the capacity of embedding space. And RotatE \cite{sun2019rotate} attempts to construct knowledge graph embedding in complex vector space. However, all of the above-mentioned KGE methods only deal with text modality triplets, our method adequately exploits the potentials of the multi-modal entity with text, video frames and audio, which means that the embedding representation of our method is more informative.

\subsection{Vision-Language Multi-Modal Pre-training}
The recent rise in pre-trained unsupervised language models like BERT \cite{devlin2018bert}, GPT-2 \cite{radford2019language}, GPT-3 \cite{brown2020language}, etc, shows that with abundant data, pre-trained models get better performance boost as model parameter capacity gets larger. Inspired by these works, similar attempts like \cite{lu2019vilbert} and \cite{tan2019lxmert} in Vision-Language Pre-training research appear. Combined with KG, we design a unified multi-modality video understanding model for text, video frames and audio integration. Our model is pre-trained on plentiful videos data from our Company short video platform. The entity representation generated by our method gains consistent improvement of performance in video inference and tag inference tasks and shows that the introduction of multi-modal embedding is beneficial.

\section{Dataset}

In this section, we begin by showing the cleaning and construction process of our video knowledge graph dataset and the information of the public knowledge graph dataset which is applied in our approach. The detailed information of the mentioned dataset is provided in Table \ref{tab2}.
\subsection{Video Knowledge Graph Dataset}
\begin{figure}[t]
\centering
\includegraphics[width=.45\textwidth]{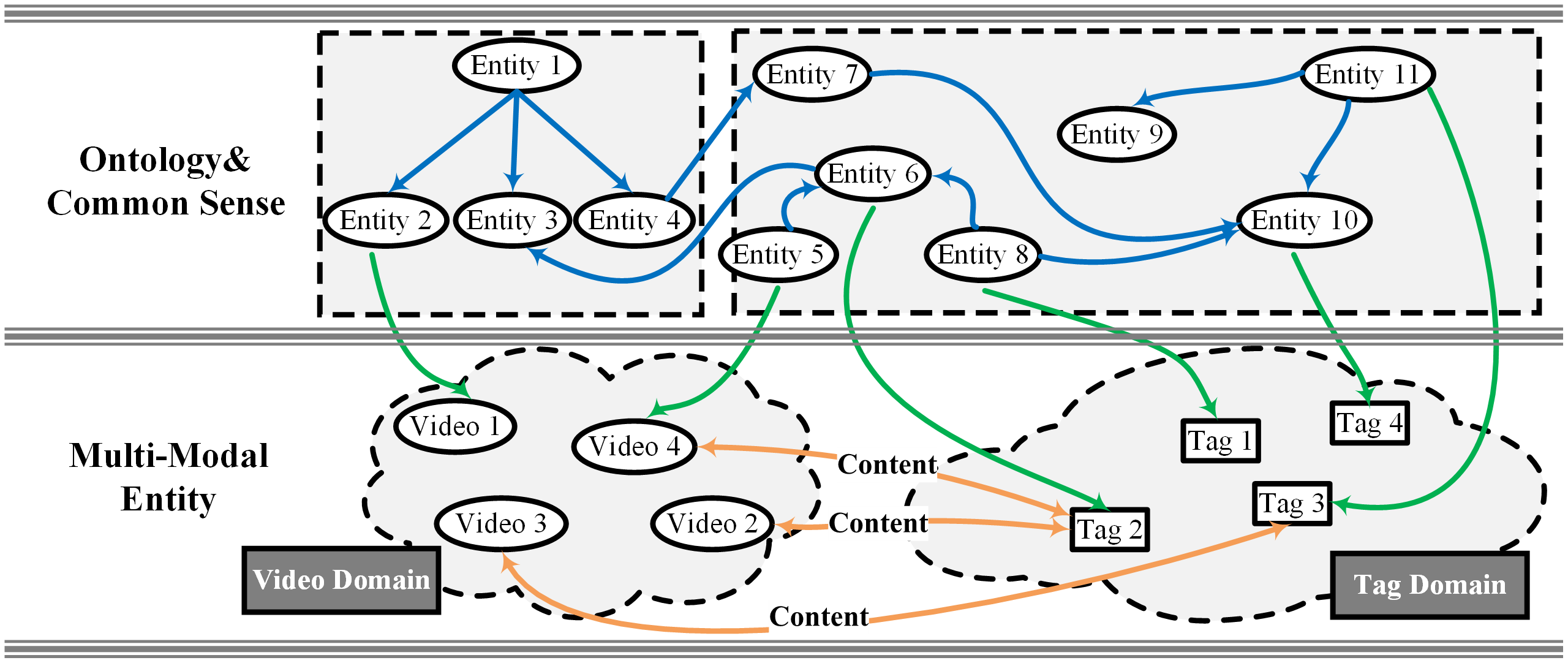}
\caption{\textbf{An overview of the proposed video knowledge graph dataset.} Our dataset is divided into two parts: The Multi-Modal Entity part and the Ontology\&Commom Sense part. In the multi-Modal Entity part, the videos are tagged with one content-related tag. In the Ontology\&Commom Sense part, there exists rich common sense and real-world knowledge triplets which comes from the Company Knowledge Graph dataset. The videos and tags are linked to the entity of knowledge graph triplets with the entity linking algorithm.}
\label{fig2}
\vspace{-0.4cm}
\end{figure}

The overview of our heterogeneous video knowledge graph dataset is shown in Figure \ref{fig2}. In this section, we will introduce multi-modal entity and linking process of videos and knowledge graph entities in brief. More details on video tagging, entity linking process and proposed datasets are presented in the supplementary material.
\subsubsection{Video \& Tag} Our video set is retrieved from Company short video platform which contains over sixty billion videos and over forty million videos are randomly sampled from the high-quality video library to form the Company-400M dataset. Based on the company short video platform, all videos uploaded to the library will be tagged by the author. We adopt the author-edited tags and filter the tags via a verification model based on pre-trained tag embeddings. In order to get the fine-grained tags, we count the video number of each tag and only reserve the tag with no more than a certain number of videos. If a video has multiple tags, we only keep the tag with the least video number. The human evaluation results for tagging accuracy is more than 90\%. 
Each video is tagged with one tag, either from the author tagging or model generation, for label. This tag represents a summary of video content and the total number of the unique tag is over 400,000 in the Company-400M dataset. The distribution of video data is balanced out during the video retrieving and tagging process.

Three types of modalities are retrieved in our method, which are $\mathcal{V}$: video frames, $\mathcal{A}$: audio and $\mathcal{T}$: text. For modality $\mathcal{V}$, four video frames are extracted from the original video averagely. For modality $\mathcal{A}$, the Automatic Speech Recognition (ASR) model is implemented for text extraction and the extracted text is added to $\mathcal{T}$ modality. $\mathcal{T}$ modality consists of four types of texts, namely, ASR text mentioned above, caption text in video frames that are captured by Optical Character Recognition (OCR) model, video title and video meta description edited by authors. These texts are then tokenized with BERT wordpiece tokenizer. Eventually, the video frames and text sequence are concatenated to form the final input of the video understanding model.

\subsubsection{Entity Linking \& Relation}
To establish the video inference tasks, every video is linked to the entity in Company Knowledge Graph dataset using a pre-trained multi-modal entity-linking model \cite{shen2014entity} \cite{gan2021multimodal}. Among the results generated by the entity linking model, we only keep the result with the highest confidence rate and the low confidence linking results that are under a certain threshold are filtered. So each video can only be linked to one entity. The entity-linking results generated by AI model is also evaluated by human with an accuracy of around 95\%. Finally, we get the Company-5M dataset. In this dataset, the number of head and tail text entities are 84,838 and 191,474 respectively. The 84,838 head entities include fine-grained entities which may be found in the specific modality of video such as an object in the video frames, the speech of background audio, or the keyword in caption and video title. For example, the head entity can be the person (artist, athlete, president), location (restaurant, mountain, city) or event (accident, holiday, sports event), etc. As for 191,474 tail entities, 13.31\% of tail entities can be found in head entities, while the rest 86.69\% of tail entities are the general concepts, taxonomy or other common sense related entities such as Nature,Film,etc. The relations which refer to the common sense pairs have a proportion of around 90\% such as isA, isPropertyOf, Co-occurrence, while the reset of relations is the factual relations such as DayOfbirth, LocatedIn, SpouseOf, etc, which have a proportion of around 10\%. And the number of unique entity and triplet is 248,324 and 832,577 while the number of relation is 5150. The number of linked videos is 5,714,531. Note that every head entity in triplet will be linked to at least one video but some of the tail entity can not be linked to any video. For those linked entities, one text entity may be linked to a number of videos but one video can only be linked to one text entity. Referring to the lengths of video and statistics of video and tags, the company's video library mainly consists of short video, so most of video length is no more than 3min.

\subsection{Public Dataset}
Owing to the lack of public video knowledge graph dataset, we elaborately design experiments on the subset of public Chinese Wikipedia knowledge dataset: CN-DBpedia \cite{xu2017cn} to demonstrate the superiority of our approach. The original dataset contains over 9,000,000 encyclopedia entity and over 67,000,000 relation triplets. Based on CN-DBpedia, we extract the matched relation triplets which are relevant to our Company knowledge graph dataset to form the CN-DBpedia sub dataset. The CN-DBpedia sub dataset contains 465,714 triplets, 101,002 entities and 4,987 relations. Then we randomly sampled 4,000 triplets from the sub dataset to form the test set while the remaining triplets constitute the train set.
\begin{figure*}[!h]
\centering
\includegraphics[width=.8\textwidth]{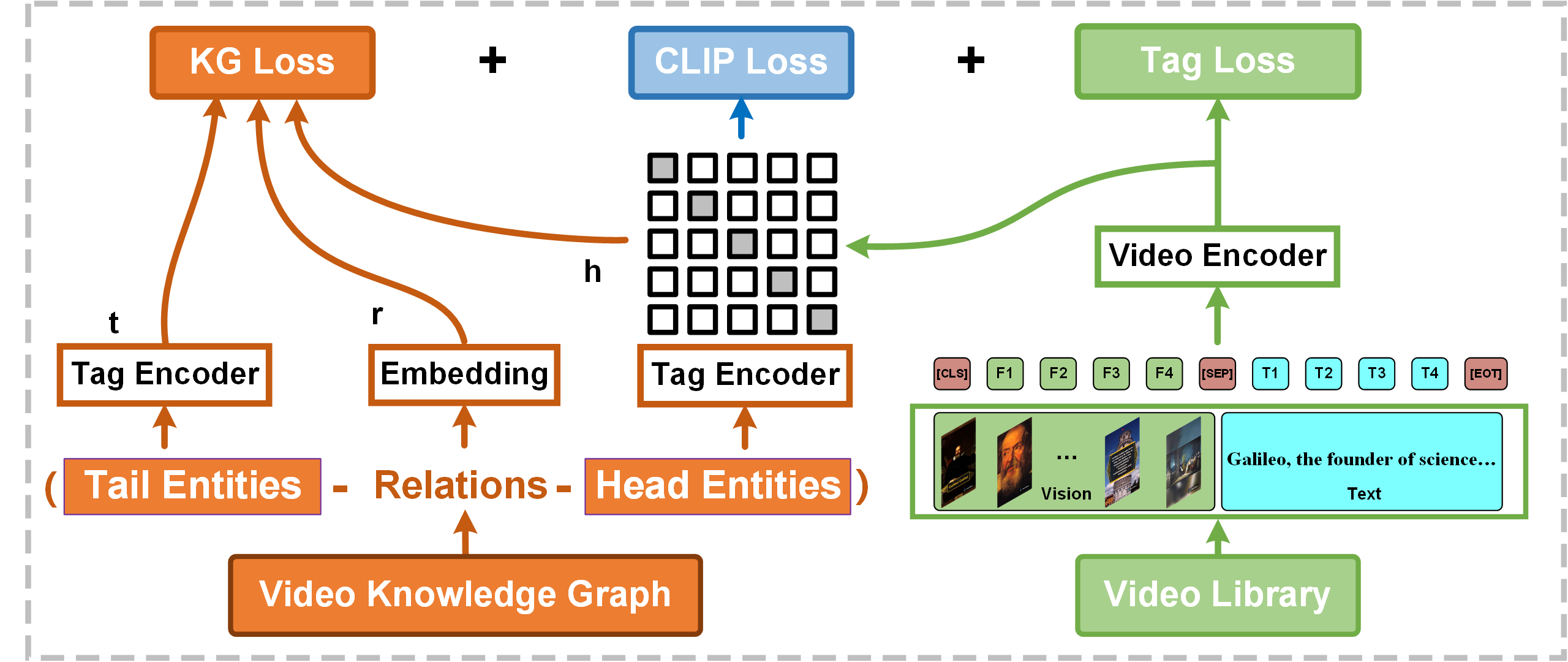}
\caption{\textbf{The proposed framework.} We take the output of video encoder as the head entity embeddings and jointly train knowledge embedding and video embedding in the same model.}
\label{fig3}
\end{figure*}

\section{Method}
In this section, we first present the content-based video understanding model which integrates different modality from video and generates embeddings representation of video. Then, we discuss the CLIP-based \cite{radford2021learning} method which associates the tag embedding into the same space of video embeddings. Lastly, we elaborate the unified architecture for the heterogeneous knowledge inference task which is depicted in Figure \ref{fig3}.
\subsection{Multi-modality Video Understanding Encoder}
Based on the video library and corresponding tag label presented in the previous section, we design a transformer-based model for video embedding extraction. Similar to many face recognition methods \cite{deng2019arcface} \cite{liu2017sphereface} \cite{wang2018cosface}, we model video understanding problem as a classification task and use the input of the classification layer as video embedding while the classification target is the tag label corresponding to the video. Our video encoder model adopts the BERT large \cite{devlin2018bert} as the backbone and it takes the tokens from three modalities of image, speech, and text as input. The video frames are tokenized by pre-trained ViT \cite{dosovitskiy2020image} with 16 patches, 768 embedding dimensions, 12 encoder layers and 12 attention heads while speech text and text data are tokenized with the the BERT wordpiece tokenizer. Note that we use ASR to convert speech audio to text. We concatenate the tokens of three modalities as the final input tokens. The tokens from different modalities are split by the \texttt{<SEP>} token. And we insert a \texttt{<CLS>} token and the \texttt{<EOT>} token at the beginning and end of the input tokens. We first project the final hidden state $C$ of the corresponding \texttt{<CLS>} token in the last encoder layer into the knowledge embedding space $\mathbb{R}^{k}$ through the projection head:
\begin{equation}
{z}_{V}=\sigma\left(C \cdot {W}^{T}_{1} + {b}\right)
\end{equation}
where $W_{1}\in{\mathbb{R}^{k \times H}}$ and $b \in {\mathbb{R}^{k}}$ are the learnable weight and bias vector, respectively. Then the video representation ${z}_{V}$ is fed to the classification layer with weight $W_{2}\in{\mathbb{R}^{T \times k}}$, where $T$ is the number of unique tag in Company-400M dataset. The scoring function for the input tokens with tag $t$ is $\mathbf{s}=\operatorname{softmax}\left({z}_{V} \cdot  W_{2}^{T}\right)$, $\mathbf{s} \in \mathbb{R}^{T}$ is a $T$ dimensional vector with $s_{i} \in[0,1]$ and $\sum_{i}^{T} s_{i}=1$. The loss for tag classification is computed by the following cross-entropy loss with $\mathbf{s}$ and tag target label $t$:

\begin{equation}
\mathcal{L}_{\mathrm{TAG}}=-\sum_{i=1}^{T} y_{i} \log \left(s_{i}\right)
\end{equation}

where $y_{i}$ is the tag classification indicator for the input tokens, $y_{i}=1$ when $i = t$ and $y_{i}=0$ when $i \neq t$.

\subsection{Knowledge Graph Embedding Integration}
In this section, we will demonstrate the way we integrate knowledge graph embedding which equips video embedding with the ability to conduct video inference tasks. The overall framework is depicted in Figure \ref{fig3}. We optimize our model in three stages. First the video encoder is trained for video understanding, then we project the video embedding into the same space of tag embedding by CLIP-based method, and finally we jointly optimize the KGE, CLIP and video understanding objectives in one model.

\noindent \textbf{Stage-One:} We first initialize the video understanding encoder $\mathbf{E_{video}}$ and it is pre-trained on the Company-400M video dataset. For every batch of videos from the video library, the video understanding encoder is optimized based on the tag classification task which is supervised by $\mathcal{L}_{\mathrm{TAG}}$. After the pre-training process, the output of the project head represents the embeddings of the video.

\noindent \textbf{Stage-Two:} Then, the tag encoder $\mathbf{E_{tag}}$ is added to encode the tag into the same semantic space of video embedding and it is pre-trained based on the CLIP task. The tag encoder is based on BERT backbone and the tag embedding is generated from the hidden state of \texttt{<CLS>} token. We train the CLIP retrieval task on the video-tag pairs $\left(V^{(i)}, T^{(i)}\right) \in {B}$ from the Company-400M video dataset, where ${B}$ is the mini-batch, $V^{(i)}$ is the video and $T^{(i)}$ is the corresponding tag. The $V^{(i)}$ and $T^{(i)}$ are fed to $\mathbf{E_{video}}$ and $\mathbf{E_{tag}}$ respectively, and the video embedding $z_{V}^{(i)}$ and tag embedding $z_{T}^{(i)}$ generated by $\mathbf{E_{video}}$ and $\mathbf{E_{tag}}$ are optimized via InfoNCE \cite{oord2018representation} loss as follows:
\begin{equation}
\begin{aligned}
\mathcal{L}_{\mathrm{CLIP}} &=\frac{1}{B} \sum_{i}^{B}-\log \frac{\exp \left(z_{V}^{(i)} \cdot z_{T}^{(i)} / \tau\right)}{\sum_{j=1}^{B} \exp \left(z_{V}^{(i)}\cdot z_{T}^{(j)} / \tau\right)} \\
&+\frac{1}{B} \sum_{i}^{B}-\log \frac{\exp \left(z_{V}^{(i)}\cdot z_{T}^{(i)} / \tau\right)}{\sum_{j=1}^{B} \exp \left(z_{V}^{(j)} \cdot z_{T}^{(i)} / \tau\right)}
\end{aligned}
\end{equation}
where $\tau$ is the learnable temperature. During the pre-training process of tag encoder, the video encoder and project head are frozen.

\noindent \textbf{Stage-Three:} After above-mentioned pre-train process, we add the knowledge graph embedding and jointly train on the Company-5M dataset. For every video in Company-5M dataset, the tag of video will be linked to a unique head entity \texttt{h} in knowledge graph and the corresponding knowledge triplet $(h, r, t)$ of this head entity will be encoded as follows:
\begin{equation}
\begin{aligned}
\mathbf{h} &=\mathbf{E_{video}}_{\texttt{<CLS>}}\left(\text{video}_{h}\right), \\
\mathbf{t} &=\mathbf{E_{tag}}_{\texttt{<{CLS}>}}\left(\text{text}_{t}\right), \\
\mathbf{r} &=\mathbf{T}_{r}
\end{aligned}
\end{equation}
\label{tagencoders}

where $\text{video}_{h}$ is the video linked to the head entity $h$ and $\text{text}_{t}$ is the text of tail entity $t$ and both of them have a special token \texttt{<CLS>} in the input sequence. The video and tail entity are encoded by the pre-trained video encoder $\mathbf{E_{video}}$ and tag encoder $\mathbf{E_{tag}}$ respectively to get the embeddings $\mathbf{h}$ and $\mathbf{t}$, while $\mathbf{T}_{r} \in \mathbb{R}^{k}$ is the relation embeddings.

We implement the loss in \cite{sun2019rotate} as the KGE objective, which uses the same negative sampling method in \cite{bordes2013translating}:
\begin{equation}
\begin{gathered}
\mathcal{L}_{\mathrm{KG}}=-\log \sigma\left(\gamma-d(\mathbf{h} +\mathbf{r}, \mathbf{t})\right) \\
-\sum_{i=1}^{n} \frac{1}{n} \log \sigma\left(d\left(\mathbf{h} +\mathbf{r}, \mathbf{t}_{\mathbf{i}}^{\prime}\right)-\gamma\right)
\end{gathered}
\end{equation}
\label{tagencoders}

where $\left(h, r, t_{i}^{\prime}\right)$ are negative samples, $\gamma$ is the margin, $\sigma$ is the sigmoid function and $d\left(\cdot\right)$ is the score function in TransE \cite{bordes2013translating},
\begin{equation}
d(\mathbf{h}+\mathbf{r}, \mathbf{t})=\|\mathbf{h}+\mathbf{r}-\mathbf{t}\|_{p}
\end{equation}

Here we take the norm $p$ as 2 and the corrupted triplets are sampled by fixing the head entity and randomly sampling a tail entity.

To jointly incorporate the factual knowledge from knowledge graph and multi-modality understanding embedding into one model, we adopt the multi-task loss as shown in Figure \ref{fig3} and Equation \ref{multiloss},
\begin{equation}
\mathcal{L}=\lambda_{1}\mathcal{L}_{\mathrm{KG}}  + \lambda_{2}\mathcal{L}_{\mathrm{CLIP}} +  \lambda_{3}\mathcal{L}_{\mathrm{TAG}}
\label{multiloss}
\end{equation}

where $\mathcal{L}_{\mathrm{KG}}$, $\mathcal{L}_{\mathrm{CLIP}}$ and $\mathcal{L}_{\mathrm{TAG}}$ are the losses for KGE, CLIP and video understanding correspondingly while $\lambda_{1}$, $\lambda_{2}$ and $\lambda_{3}$ are the tradeoff parameters. During the joint training process on the Company-5M dataset, the CLIP loss is open for further optimization of the heterogeneous embedding space, which guarantees the isomorphism of video and tag embedding space. $\mathcal{L}_{\mathrm{KG}}$ implicitly integrates the knowledge from the KG dataset while the $\mathcal{L}_{\mathrm{TAG}}$ reserves the ability of video encoder for multi-modality understanding. 
\begin{table*}[htbp]
\begin{center}
\caption{The baselines and variants of our method. $\mathcal{L}_{\mathrm{KG}}$ represents the corresponding KGE loss for TransE, TransH or TransR.}
\label{tab1}
\begin{tabular}{p{2.2cm}<{\raggedright}|p{1.35cm}<{\centering}|p{1.35cm}<{\centering}|p{1.35cm}<{\centering}|p{1.35cm}<{\centering}|p{1.35cm}<{\centering}|p{1.35cm}<{\centering}|p{1.35cm}<{\centering}|p{1.35cm}<{\centering}}
\hline
Baseline    & VRV & VRT & TRT & VT & TV & $\mathcal{L}_{\mathrm{TAG}}$ & $\mathcal{L}_{\mathrm{CLIP}}$ &  $\mathcal{L}_{\mathrm{KG}}$ \\ \hline
\textbf{TransE}      & -   & -   & $\surd$   & -  & - & - & - & $\surd$ \\
\textbf{TransH}      & -   & -   & $\surd$   & -  & - & - & - & $\surd$  \\
\textbf{TransR}      & -   & -   & $\surd$   & -  & - & - & - & $\surd$ \\
\textbf{CLIP}        & -   & -   & -   & $\surd$  & $\surd$ & $\surd$ & $\surd$ & - \\
\textbf{CLIP+TransE} & $\surd$   & $\surd$   & $\surd$   & $\surd$  & $\surd$ & $\surd$ & $\surd$ & $\surd$  \\
\textbf{CLIP+TransH} & $\surd$   & $\surd$   & $\surd$   & $\surd$  & $\surd$ & $\surd$ & $\surd$ & $\surd$ \\
\textbf{CLIP+TransR} & $\surd$   & $\surd$   & $\surd$   & $\surd$  & $\surd$ & $\surd$ & $\surd$ & $\surd$ \\
\textbf{Ours}        & $\surd$   & $\surd$   & $\surd$   & $\surd$  & $\surd$ & $\surd$ & $\surd$ & $\surd$ \\ \hline
\end{tabular}
\end{center}
\end{table*}
\section{Experiments}
We conduct comprehensive experiments to evaluate our method in Tag-to-Video (TV) retrieval task, Video-to-Tag (VT) retrieval task, Tag-Relation-Tag (TRT) inference task, Video-Relation-Tag (VRT) inference task and Video-Relation-Video (VRV) inference task. Apart from that, due to the lack of public video knowledge dataset, we compare our method on the CN-DBpedia sub dataset with three classic knowledge graph embedding methods which are TransE, TransH and TransR following the standard implementation by \cite{han2018openke}. We implement all experiments on a machine with 1,000GB memory and 8 A100 GPUs. 
\subsection{Implementation Detail}
\textbf{Baseline:} Due to the lack of methods with the same objective on video inference, we design several classic embedding-based baselines and variants of our method. To be specific, they are: \textbf{TransE}: The classic KGE method which only deals with the TRT task, \textbf{TransH}: The KGE method based on relation-specific entity embeddings which only deals with the TRT task, too. \textbf{TransR}: An improved method which gains more performance boosting compared with TransH while only capable of dealing with TRT task. \textbf{CLIP}: A variant of our method without integration of knowledge graph embedding and is only pre-trained on Company video dataset with $\mathcal{L}_{\mathrm{TAG}}$ and $\mathcal{L}_{\mathrm{CLIP}}$. This baseline can only deal with VT and TV retrieval tasks. \textbf{CLIP+TransE}: A two-stage method for video inference. Given a video and relation, this method will get the head entity of the video by looking in the Company-5M dataset then conduct the pure text induction based on the TransE method to get the tail entity text. Finally, it retrieves the videos using the CLIP-based video embedding and tag embedding. This method is capable of conducting all the above-mentioned tasks. \textbf{CLIP+TransH}: Similar to CLIP+TransE, but the text induction method is replaced by TransH. \textbf{CLIP+TransR}: Similar to CLIP+TransE, but the text induction method is replaced by TransR. \textbf{Ours}: Our proposed method is capable of conducting all the above-mentioned tasks. All baselines and their corresponding abilities are shown in Table \ref{tab1}.

\textbf{Setup Detail:} During the process of training and inference of our proposed method, the dimension of relation embedding space $k^{(\mathbb{R})}$=128, which is the same as the dimension of output embeddings from the video encoder and tag encoder. The learnable temperature $\tau$ in the InfoNCE loss is initialized as 0.07. The margin $\gamma$ in KG loss is set as 4 and the negative samples number is 5. The transformer setup of the video encoder and tag encoder is the same: the number of transformer layers is 24, the hidden size is 1024, the number of attention heads is 16 and the sequence length is 128. And the tradeoff parameters $\lambda_{1}$, $\lambda_{2}$ and $\lambda_{3}$ are set to be 0.35, 0.35 and 0.3 respectively. We use Adam \cite{loshchilov2017decoupled} to optimize most of the training process with learning rate=$10^{-4}$ and global batchsize=512, but during the CLIP pre-train process in Stage-Two the global batchsize is set to $880$ for convergence efficiency. The retrieval process in the following experiments is all based on the cosine similarity between video or text embeddings. Two widely used metrics are adopted to evaluate the performance of the model which are Mean Rank (MR) and HITS@n.

\begin{table*}[t]
\begin{center}
\caption{The performance comparison of VT and TV retrieval task.}
\label{VTandTV}
\begin{tabular}{p{0.8cm}<{\raggedright}|p{1.5cm}<{\centering}p{1.5cm}<{\centering}p{1.5cm}<{\centering}p{1.5cm}<{\centering}|p{1.5cm}<{\centering}p{1.5cm}<{\centering}p{1.5cm}<{\centering}p{1.5cm}<{\centering}}
\hline
\multicolumn{1}{c|}{\multirow{2}{*}{Method}} & \multicolumn{4}{c}{VT}                 & \multicolumn{4}{|c}{TV}                  \\ 
\multicolumn{1}{c|}{}                        & MR         & HITS@1 & HITS@3 & HITS@10 & MR          & HITS@1 & HITS@3 & HITS@10 \\ \cline{1-9}
CLIP                                        & 14515.2419 & 0.0885 & 0.1487 & 0.2252  & 12038.8518 & 0.1143 & 0.1864 & 0.2660  \\  
Ours     & \textbf{10622.3440} & \textbf{0.1241} & \textbf{0.2186} & \textbf{0.3438}  & \textbf{9030.5341} & \textbf{0.2786} & \textbf{0.3907} & \textbf{0.4759}
\\ \cline{1-9}
\end{tabular}
\end{center}
\end{table*}

\begin{table*}[h!]
\begin{center}
\caption{The performance comparison of TRT inference task.}
\begin{tabular}{p{2.2cm}<{\raggedright}|p{1.35cm}<{\centering}p{1.35cm}<{\centering}|p{1.35cm}<{\centering}p{1.35cm}<{\centering}|p{1.35cm}<{\centering}p{1.35cm}<{\centering}|p{1.35cm}<{\centering}p{1.35cm}<{\centering}}
\hline
\multirow{2}{*}{Method} & \multicolumn{2}{c|}{MR} & \multicolumn{2}{c|}{HITS@1} & \multicolumn{2}{c|}{HITS@3} & \multicolumn{2}{c}{HITS@10} \\
                        & head        & tail     & head         & tail        & head         & tail        & head         & tail         \\ \hline
TransE(/+CLIP)           & 1619.8007   & 3.0365   & 0.2727       & 0.5510      & 0.4835       & 0.7090      & 0.6622       & 0.9650       \\
TransH(/+CLIP)           & 1466.6250   & 2.9182   & 0.2700       & 0.5472      & 0.4895       & 0.7210      & 0.6675       & 0.9702       \\
TransR(/+CLIP)           & 1222.8027   & 2.3595   & 0.2712       & 0.5522      & 0.5322       & \textbf{0.7892}      & 0.6982       & \textbf{0.9920}   \\ 
Ours(+Embed)                    & \textbf{1151.6250}          & \textbf{2.2046}     & \textbf{0.3250}      & \textbf{0.6600}       & \textbf{0.5750}      & 0.7575      & \textbf{0.7650}      & 0.9850  \\ \hline  
\end{tabular}
\label{TRTinf}
\end{center}
\end{table*}

\begin{table*}[h!]
\begin{center}
\caption{The performance comparison of VRV and VRT inference task.}
\label{VRVandVRT}
\begin{tabular}{p{2cm}<{\raggedright}|p{1.35cm}<{\centering}p{1.35cm}<{\centering}p{1.35cm}<{\centering}p{1.35cm}<{\centering}|p{1.35cm}<{\centering}p{1.35cm}<{\centering}p{1.35cm}<{\centering}p{1.35cm}<{\centering}p{1.35cm}<{\centering}}
\hline
\multirow{2}{*}{Method} & \multicolumn{4}{c|}{VRV}                & \multicolumn{4}{c}{VRT}             \\ 
                        & MR         & HITS@1 & HITS@3 & HITS@10 & MR      & HITS@1 & HITS@3 & HITS@10 \\ \hline
CLIP+TransE             & 23356.6640 & 0.0340 & 0.0618 & 0.0961  & 52.3991 & 0.0508 & 0.1019 & 0.3981  \\
CLIP+TransH            & 23168.7382 & 0.0368 & 0.0674 & 0.1063  & 34.7941 & 0.0498 & 0.1198 & 0.4506  \\
CLIP+TransR             & 27608.6244 & 0.0475 & 0.0884 & 0.1396  & 25.5660 & 0.0508 & 0.2152 & 0.5869  \\
Ours                    & \textbf{8357.8196} & \textbf{0.2759}      & \textbf{0.3977}       & \textbf{0.5632}        & \textbf{13.4505} & \textbf{0.1144}   & \textbf{0.4308}       & \textbf{0.7642}  \\ \hline     
\end{tabular}
\end{center}
\end{table*}
\begin{table*}[h!]
\begin{center}
\caption{Comparison of head and tail entity prediction result on CN-DBpedia sub dataset.}
\begin{tabular}{p{2.2cm}<{\raggedright}|p{1.35cm}<{\centering}p{1.35cm}<{\centering}|p{1.35cm}<{\centering}p{1.35cm}<{\centering}|p{1.35cm}<{\centering}p{1.35cm}<{\centering}|p{1.35cm}<{\centering}p{1.35cm}<{\centering}}
\hline
\multirow{2}{*}{Method} & \multicolumn{2}{c|}{MR} & \multicolumn{2}{c|}{HITS@1} & \multicolumn{2}{c|}{HITS@3} & \multicolumn{2}{c}{HITS@10} \\
                             & head       & tail      & head         & tail        & head         & tail        & head         & tail         \\ \hline
TransE                       & 10722.1680 & 2059.1123 & 0.1795       & 0.2720      & 0.1860       & 0.5050      & 0.1970       & 0.7268       \\
TransE+Embed                   & \textbf{9556.5312}          & \textbf{1956.0928}         & \textbf{0.2885}            & \textbf{0.3377}           & \textbf{0.2932}            &   \textbf{0.5650}         & \textbf{0.3896}            & \textbf{0.8750}            \\ \hline
TransH                       & 11189.0244 & 2349.0303 & 0.1780       & 0.2448      & 0.1838       & 0.4878      & 0.1908       & 0.7288       \\
TransH+Embed                   & \textbf{9432.7176}          & \textbf{1839.1133}         & \textbf{0.2845}            & \textbf{0.3557}           & \textbf{0.2981}            & \textbf{0.5481}           & \textbf{0.3875}            & \textbf{0.8725}            \\ \hline
TransR                       & 13613.5000 & 2742.9885 & 0.1778       & 0.2440      & 0.1858       & 0.4173      & 0.1948       & 0.6303       \\
TransR+Embed                   & \textbf{9628.6962}          & \textbf{2075.3750}         & \textbf{0.2888}           & \textbf{0.3575}           & \textbf{0.2975}            & \textbf{0.4725}           & \textbf{0.3817}            & \textbf{0.7125}            \\ \hline 
\end{tabular}
\label{CNDBpedia}
\end{center}
\end{table*}

\subsection{Content based Retrieval Task Performance}
 The primary role of the CLIP pre-training process is to guarantee the isomorphism of video and tag embedding. This isomorphism is a contributing factor in the performance of Video-to-Tag and Tag-to-Video retrieval task. Since the tag can be regarded as the summary of the video, embedding of related tags and videos should be clustered. In accordance with the above hypothesis, we test the \textbf{CLIP} and \textbf{Ours} baselines on Company-5M dataset. The Company-5M dataset is divided into train set and test set separately and the video proportion in train set and test set is 90:5. The performance result on test set is shown in Table \ref{VTandTV}.
 
As shown in Table \ref{VTandTV}, our method gains 11.86\% and 20.99\% improvement in TV and TV task respectively on HITS@10 than the baseline CLIP. The results show that when jointly integrating the knowledge graph embedding loss $\mathcal{L}_{\mathrm{KG}}$ with $\mathcal{L}_{\mathrm{CLIP}}$ and $\mathcal{L}_{\mathrm{TAG}}$, the performance of these tasks gains a significant margin of improvement which is likely due to the information gained from the external factual knowledge. This result shows that the participation of knowledge graph embedding benefits the content retrieval task. However, as shown in Table \ref{VTandTV}, there is a large performance difference between VT and TV tasks, where the performance of TV is better than VT. This phenomenon
 may be caused by the imbalance of video and tag proportions. The ratio of video and entity is 23:1 in Company-5M dataset, meaning that there are more candidates in Tag-to-Video task than Video-to-Tag task since each video is tagged with one tag while each tag may be related to numerous videos during the generation process of Company-5M dataset. Thus the performance difference is reasonable.

\subsection{TRT Inference Task Performance}
In this experiment, we compare the performance of the baselines that are capable of the TRT inference task and the result on test set is shown in Table \ref{TRTinf}. These baselines are tested on the pure text part of our Company-5M dataset and the train/test set split is the same in the TV/VT experiment. Owing to the single text modality in this experiment, the TransE/TransH/TransR+CLIP baselines degrade into TransE/TransH/TransR baselines, so the corresponding test result is merged in Table \ref{TRTinf}. In this experiment, when it comes to our methods, the head entity and tail entity text embedding is extracted by the pre-trained tag encoder and only the relation embedding is trained.

As shown in Table \ref{TRTinf}, our method obtains 5.23\% and 10.78\% improvement in head entity and tail entity prediction task respectively on HITS@1 which outperforms all pure KGE methods. This result indicates that the tag encoder not only generates better entity representations but also acquires rich multi-modal information from the video encoder. This result also indicates that compared to the low-dimensional embedding from the pure KGE method, the entity embedding representation coming from our tag encoder is more informative and beneficial when dealing with the pure text entity prediction tasks.

\subsection{VRV and VRT Inference Task Performance}
In this experiment, we report the result of Video-Relation-Video(VRV) and Video-Relation-Tag(VRT) inference task. The definition of VRV task is that when given a video and relation, the induction algorithm should retrieve the relevant videos while the VRT task is acquiring the relevant tag. This experiment is conducted on our Company-5M test set. Referring to the ground truth of VRV and VRT task, every video in Company-5M dataset is linked to a knowledge graph head entity and the relations associated with this entity will be selected as the given relations, while the tail entity and its linked videos will serve as the ground truth of VRT and VRV task. We compare our method with the two-stage methods (CLIP+TransE/TransH/TransR) which are explained in the previous section.

As shown in Table \ref{VRVandVRT}, our method obtains a large margin of 42.36\% and 17.73\% improvement in VRV and VRT task respectively on HITS@10 which outperforms all two-stage KGE-based methods. The reason why of low performance of these two-stage methods is mainly the lack of synthetical integration of multi-modality entity and knowledge graph embedding. In essence, these two-stage methods are limited by the individual performance of pure text KGE stage and CLIP retrieval stage. Take CLIP+TransE baseline in VRV task as an example, during pure text induction, if tail entity predicted by TransE is wrong, the recalled videos will be totally different which may lead to extreme performance degradation. And this is also the reason why the performance of VRT task is much better than VRV task.

\subsection{Generality}
In this section, we report the generalization performance on the existing public knowledge graph dataset. Owing to the lack of video knowledge graph dataset, we conduct our experiments on CN-DBpedia sub dataset. On this sub dataset, we randomly sample 4,000 triplets to form the test set while the remaining 461,714 triplets serve as train set. Since there is only text modality in this dataset, we extract the entity embedding from the pre-trained tag encoder and concatenate the extracted embedding with the KGE entity embedding to test our method. The fusion embedding will be fed to a fully connected layer for dimensionality reduction. The final dimension of fusion entity embedding is the same as the relation embedding.
As is shown in Table \ref{CNDBpedia}, when the extracted embedding is concatenated to the pure KGE methods (TransE, TransH, TransR), the performance of all of these methods improves. This shows that the embedding from the tag encoder is meaningful and the tag embedding implies extra knowledge learned from the multi-modal video entity, proving the effectiveness of our methods again.

\subsection{Feature Representation Visualization}
\begin{figure}[!h]
\centering
\includegraphics[width=.45\textwidth]{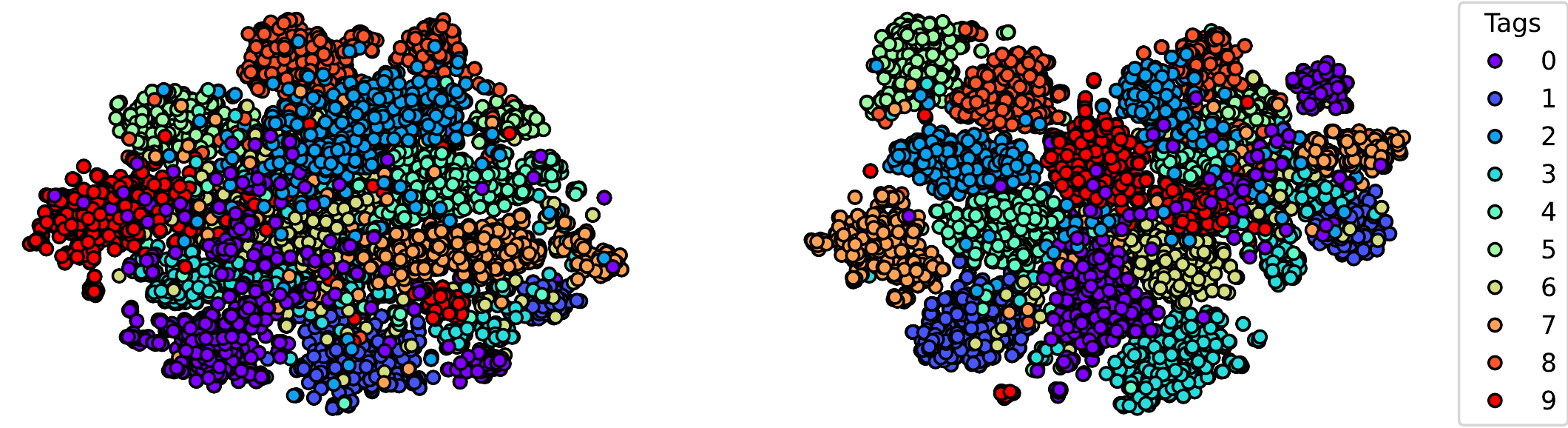}
\caption{\textbf{Visualization result of the video embedding on Company-5M dataset.} The left image represents the result of baseline \textbf{CLIP} while the right one represents the result of our method.}
\label{tnse}
\end{figure}

To further investigate the video embedding distribution generated by our model, we visualize the video embedding feature of baseline \textbf{CLIP} (left) and \textbf{Ours} (right) on the Company-5M dataset in Figure \ref{tnse} using t-SNE \cite{van2008visualizing}. Here, ten tags are chosen from the Company-5M dataset and we randomly sample 1000 videos for each tag. In Figure \ref{tnse} each point stands for the video embedding and the color represents the tag label of the corresponding video. Obviously, our model has more separate embedding clusters, especially on the tag 5, 7 and 8.

The visualization result reflects that when jointly training $\mathcal{L}_{\mathrm{TAG}}$, $\mathcal{L}_{\mathrm{CLIP}}$ and $\mathcal{L}_{\mathrm{KG}}$ on the Company-5M dataset, the video and tag embedding cluster better compared with the baseline \textbf{CLIP} which only trains with $\mathcal{L}_{\mathrm{TAG}}$ and $\mathcal{L}_{\mathrm{CLIP}}$. The supervision of knowledge embedding space helps the embedding of tag and video cluster better, showing that the external factual knowledge does benefit the VT and TV content-based retrieval task.
 
\section{Conclusion}
In this paper, we form a heterogeneous dataset that allows for multiple inference tasks including Tag-Relation-Tag, Video-Relation-Video and Video-Relation-Tag. Based on this dataset, we design a unified model for video understanding representation and knowledge graph embedding integration. We train our model with CLIP, Knowledge Graph Embedding and video understanding objectives all together to align the video representation and knowledge embedding into the same semantic space. Comprehensive experiments on Company and public dataset demonstrate the effectiveness of our methods both in video understanding and on video inference tasks. In the future, we will conduct more experiments on public multi-modal knowledge graph datasets such as FB15K237 \cite{toutanova2015observed} and investigate advanced methods for integrating the two semantic space.

\begin{acks}
This research was supported by the National Key Research and Development Program of China under Grant No. 2018AAA0100400, and the National Natural Science Foundation of China under Grants 61802407, 61976208 and 62071466.
\end{acks}


\appendix

\section{Ethics, Data, And Privacy}
We firmly affirm that we adhere to the most stringent standards for complying with regulations and ethical codes relating to data safety and privacy preservation. To minimize any potential risks and issues, we have implemented the following measures:

\textbf{Data Desensitization.} All videos in our dataset are sourced from public video-sharing platforms, and we have taken steps to remove any personal information that could potentially identify the video owners. This includes, but is not limited to, identifiers, meta-data, addresses, and profiles.

\textbf{Data Storage.} We are authorized to store and maintain the entire dataset to support future research. For ease of use, the dataset is comprised of video URLs and linked KG triplets, and we will check video accessibility before releasing the data.

\textbf{User Privacy.} We value user privacy highly and have taken measures to ensure that all videos in the dataset are publicly available on video-sharing platforms with the consent of the users for academic research and public use. All sensitive information about users has been removed from the dataset.

\textbf{License.} The dataset is for non-commercial use only and requires a signed agreement before use.

\textbf{Annotator Related.} Our annotators are highly skilled and experienced in preparing data, and have received instruction on how to avoid potential risks and dangers during the annotation process. They are fairly compensated in accordance with local laws.

\section{Video Visualization}
\begin{figure}[!h]
\centering
\includegraphics[width=.45\textwidth]{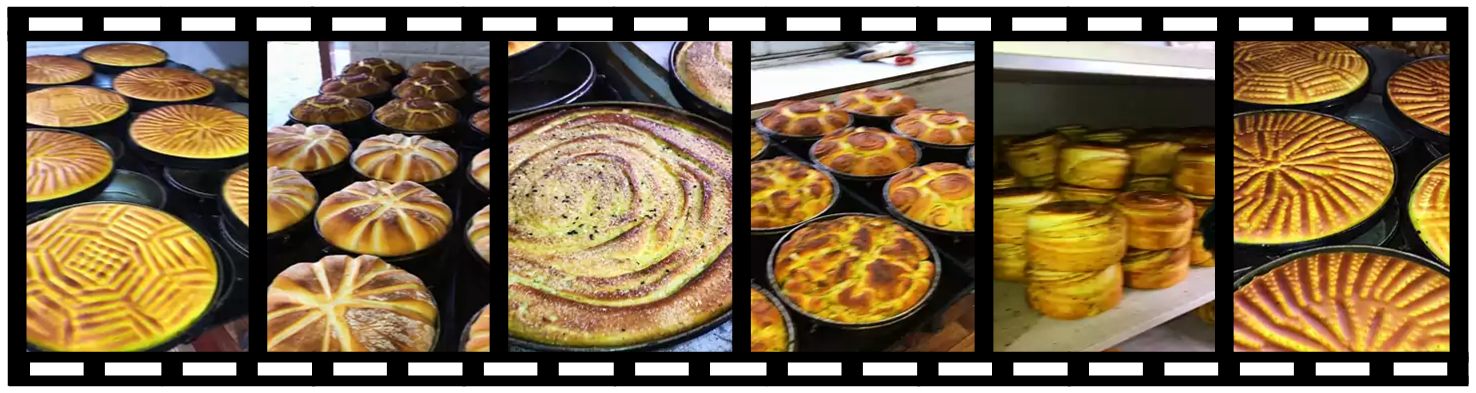}
\caption{\textbf{Visualization of the \texttt{video\_1} with tag ``bake".}}
\label{tnse1}
\vspace{-0.5cm}
\end{figure}

\begin{figure}[!h]
\centering
\includegraphics[width=.45\textwidth]{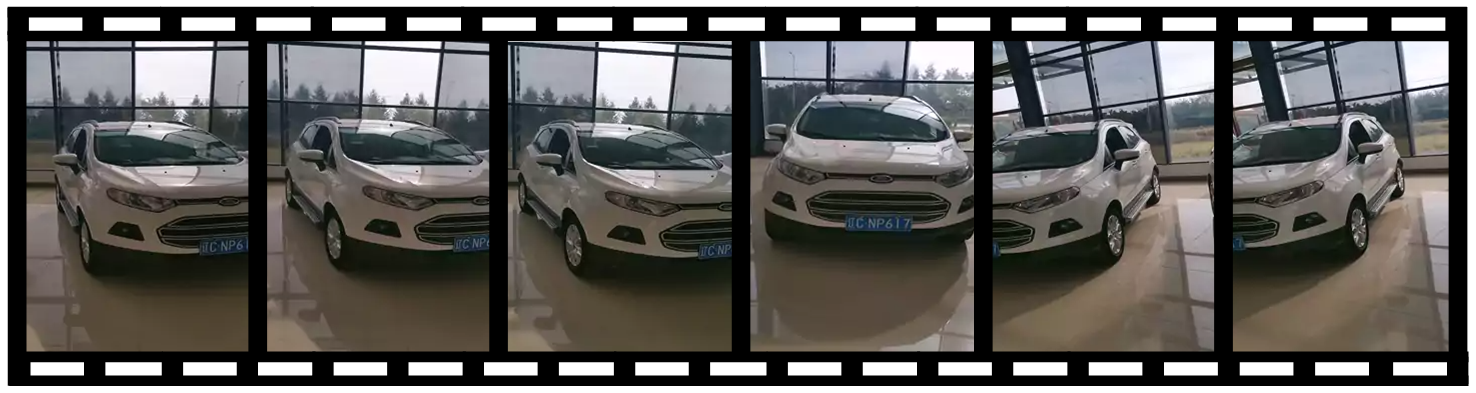}
\caption{\textbf{Visualization of the \texttt{video\_2} with tag ``car".}}
\label{tnse2}
\vspace{-0.2cm}
\end{figure}

\begin{figure}[!h]
\centering
\includegraphics[width=.45\textwidth]{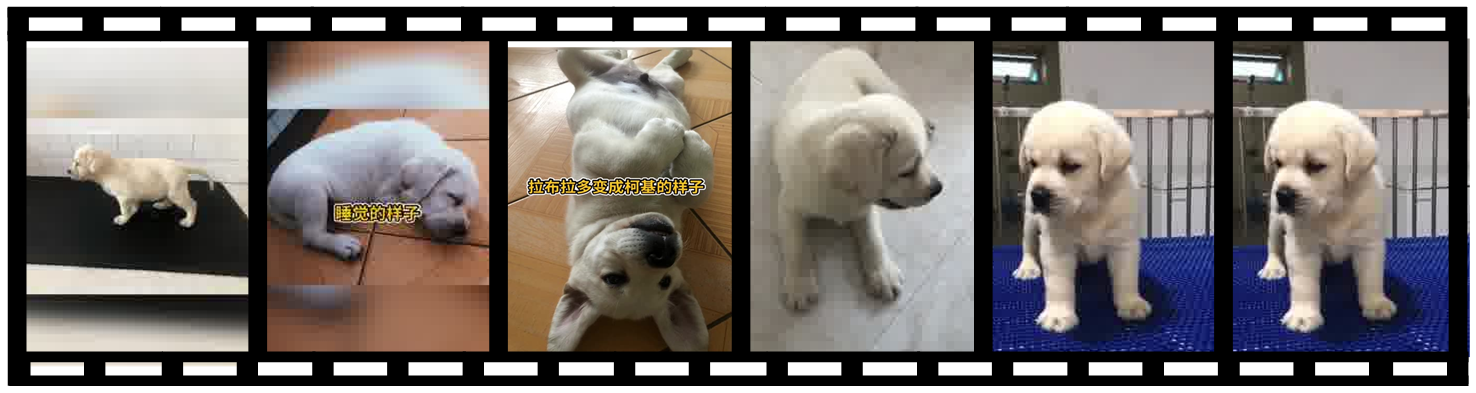}
\caption{\textbf{Visualization of the \texttt{video\_3} with tag ``dog".}}
\label{tnse3}
\vspace{-0.2cm}
\end{figure}

\begin{figure}[!h]
\centering
\includegraphics[width=.45\textwidth]{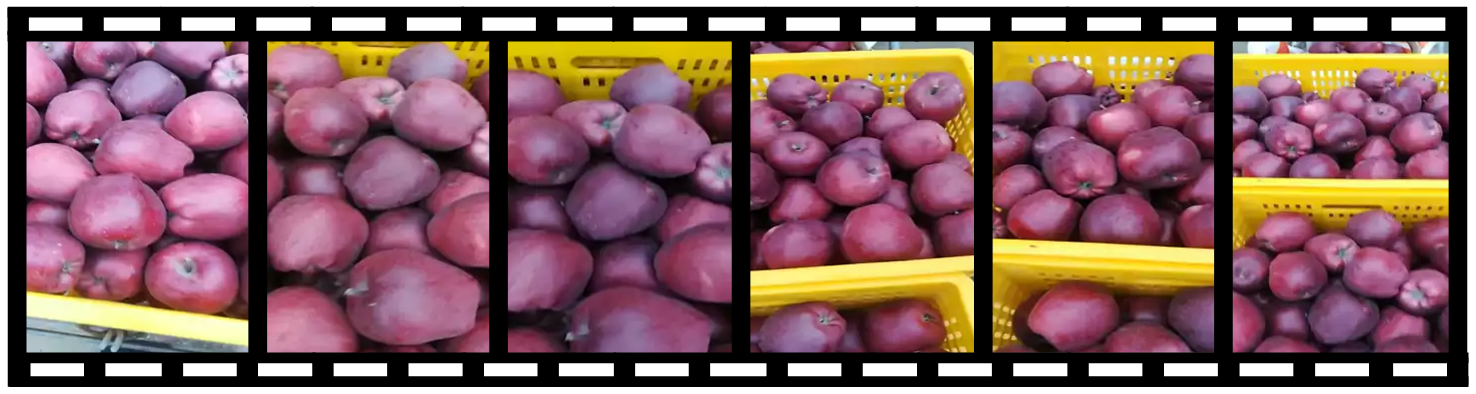}
\caption{\textbf{Visualization of the \texttt{video\_4} with tag ``apple".}}
\label{tnse4}
\vspace{-0.2cm}
\end{figure}

\begin{figure}[!h]
\centering
\includegraphics[width=.45\textwidth]{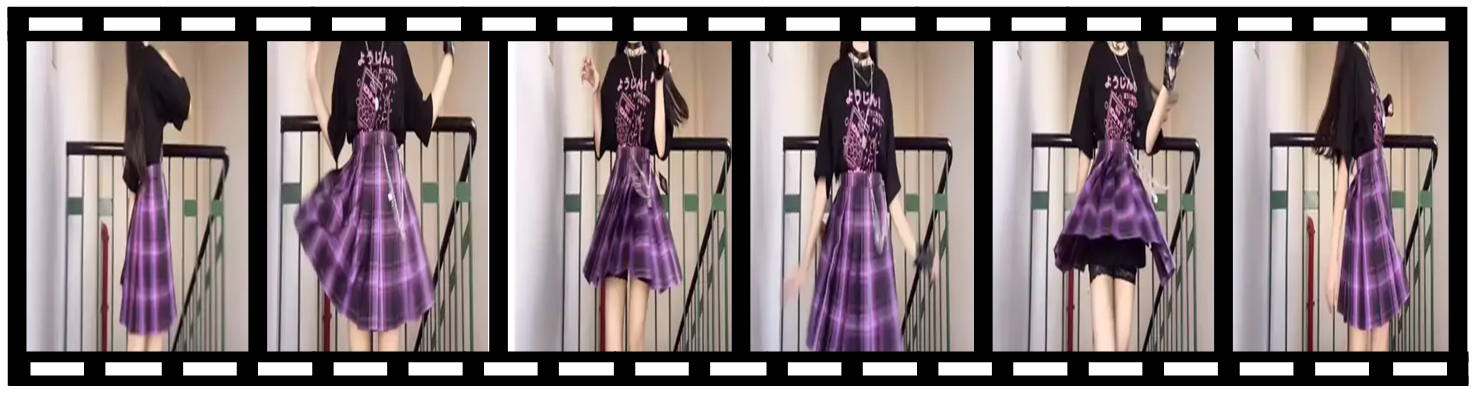}
\caption{\textbf{Visualization of the \texttt{video\_5} with tag ``JK".}}
\label{tnse5}
\vspace{-0.2cm}
\end{figure}

\begin{figure}[!h]
\centering
\includegraphics[width=.45\textwidth]{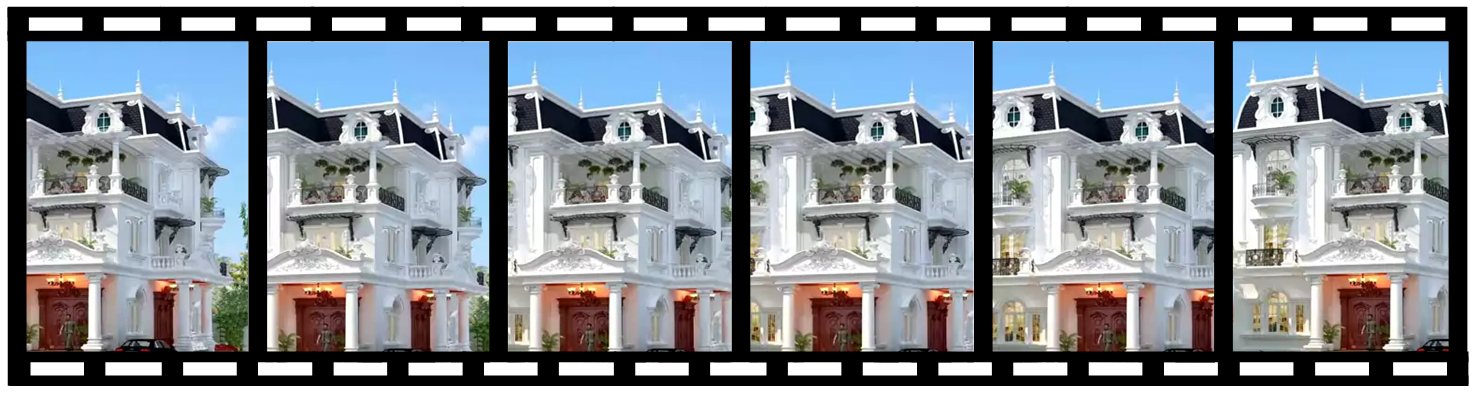}
\caption{\textbf{Visualization of the \texttt{video\_6} with tag ``house".}}
\label{tnse6}
\vspace{-0.2cm}
\end{figure}

Due to the space limits, we randomly selected videos \texttt{video\_1}
$\cdots $\texttt{video\_6} from our dataset along with tag in Figure \ref{tnse1} to \ref{tnse6}.

\section{Release Agreement}
One must sign the following agreement in order to be permitted to use our dataset.

\begin{framed}
\begin{center}
\textbf{Dataset Usage Agreement}
\end{center}

$\bullet$ Non-Commercial Use Only: The use of our dataset is limited to non-commercial purposes, including academic research and education. Profitable or infringing activities, such as advertising, selling, or face-based applications, are strictly prohibited.

$\bullet$ No Bio-Info Analysis: You promise not to conduct any analysis with respect to bio-info in the dataset, including but not limited to faces, gender, etc.

$\bullet$ Responsibility for Usage: You acknowledge that you are fully responsible for your usage of the dataset. If you engage in any illegal behavior or cause negative influences, we reserve the right to ask you to stop immediately and eliminate the impact.

$\bullet$ Right to Terminate: In the event of any violation of this agreement, we reserve the right to stop your usage of the dataset and delete any copies.

$\bullet$ Purpose and Potential Effects: You must clearly state the purpose and potential effects of using our dataset.
\begin{center}
\textbf{Name:\underline{\quad}}\textbf{Email:\underline{\quad}}\textbf{Organization:\underline{\quad}}\textbf{Date:\underline{\quad}}\textbf{Signature:\underline{\quad}}
\end{center}

\end{framed}








\end{document}